\documentclass[authoryear,3p]{elsarticle}
\usepackage{lipsum}
\usepackage{lineno}
\usepackage{amsmath,amssymb,dsfont}
\usepackage{epsf,graphicx,psfrag}
\usepackage{wrapfig,lipsum,booktabs}
\usepackage[colorlinks]{hyperref}

\usepackage{multirow}
\usepackage{longtable,lscape}
\usepackage{cases}
\usepackage{float}
\usepackage{url}
\usepackage{caption}
\usepackage{color}
\usepackage{amsthm}
\usepackage{diagbox}
\usepackage{algpseudocode}
\usepackage[ruled,vlined,titlenotnumbered]{algorithm2e}

\SetCommentSty{mycommfont}
\SetKwInput{KwInput}{Input}                
\SetKwInput{KwOutput}{Output}
\usepackage{makecell}

\usepackage{enumitem}

\usepackage[thinc]{esdiff}
\usepackage{fancyhdr}
\usepackage[utf8]{inputenc}
\usepackage{nomencl}
\makenomenclature

\usepackage{etoolbox}
\renewcommand\nomgroup[1]{%
  \item[\bfseries
  \ifstrequal{#1}{A}{Lower Level}{%
  \ifstrequal{#1}{B}{Upper Level}{%
  \ifstrequal{#1}{C}{Other Notations}{}}}%
]}

\theoremstyle {plain}
\theoremstyle{definition}

\usepackage{tikz}
\usetikzlibrary{shapes,arrows}

\usepackage{subcaption}

\newenvironment{frcseries}{\fontfamily{frc}\selectfont}{}

\newcommand{\acknowledgments}[1]{
\vspace{6pt}\noindent{\fontsize{9}{9}\selectfont\textbf{Acknowledgments:} {#1}\par}}

\numberwithin{equation}{section}

\begin{document}

\begin{frontmatter} 


\noindent 
\textcolor{blue}{Published in: \href{https://doi.org/10.3390/app10062046}{{\color{blue} \textit{Applied Sciences} 10(6) 2046, 2020}}.
Please cite this paper as: \\Gu, Z., Li, Z., Di, X., Shi, R., 2020. An LSTM-based autonomous driving model using a Waymo Open Dataset.
\textit{Applied Sciences} 10(6) 2046.
DOI: \href{https://doi.org/10.3390/app10062046}{{\color{blue} 10.3390/app10062046}}}

\title{An LSTM-Based Autonomous Driving Model Using Waymo Open Dataset}

\date{\today}

\author[cu2]{Zhicheng Gu}
\author[cu2]{Zhihao Li}
\author[cu1,dsi]{Xuan Di}
\author[cu1]{Rongye Shi\corref{cor1}}
\ead{rongye.shi@columbia.edu}


\cortext[cor1]{Corresponding author.}
\address[cu2]{Department of Computer Science, Columbia University}
\address[cu1]{Department of Civil Engineering and Engineering Mechanics, Columbia University}
\address[dsi]{Data Science Institute, Columbia University}

\begin{abstract}
The Waymo Open Dataset has been released recently, providing a platform to crowdsource some fundamental challenges for automated vehicles (AVs), such as 3D detection and tracking. While the dataset provides a large amount of high-quality and multi-source driving information, people in academia are more interested in the underlying driving policy programmed in Waymo self-driving cars, which is inaccessible due to AV manufacturers' proprietary protection. Accordingly, academic researchers have to make various assumptions to implement AV components in their models or simulations, which may not represent the realistic interactions in real-world traffic. Thus, this paper introduces an approach to learn an long short-term memory (LSTM)-based model for imitating the behavior of Waymo's self-driving model. The proposed model has been evaluated based on Mean Absolute Error (MAE). The experimental results show that our model outperforms several baseline models in driving action prediction. Also, a visualization tool is presented for verifying the performance of the model.
\end{abstract}


\end{frontmatter}

\section{Introduction}

Autonomous vehicles (AVs) have attracted a significant amount of interest in recent years, and many leading companies, such as Waymo and Lyft, have invested a huge amount of money, manpower and engineering capabilities in developing such systems. Designing policies for an autonomous driving system is particularly challenging due to demanding performance requirements in terms of both making safe operational decisions and fast processing in real-time. Despite recent advancements in technology, such systems are still largely under exploration with many fundamental challenges unsolved. 

\subsection{Motivation}

Reliable driving policies play a crucial role in developing an effective AV system, which is also one fundamental challenge faced by researchers. No documentation reveals how the existing AV test fleets on public roads are actually programmed to drive and interact with other road users due to AV manufacturers' proprietary protection, which forms a barrier for outsiders to develop AV policies for follow-up studies. Accordingly, academic researchers have to make various assumptions to implement AV components in their models or simulations, which may not represent the realistic interactions in the simulation-based research.

Waymo’s recent release of a rich and diverse autonomous driving dataset\cite{waymo_open_dataset, sun2019scalability} attracted the attention in academia in terms of pattern recognition, 3D detection and tracking. However, researchers might be more interested in inaccessible driving policies of Waymo cars that generates the dataset. Combining with the idea of behavioral cloning and the large Waymo dataset, the lack-of-practical-policy issue can be properly addressed. To this end, this paper proposes a novel long short-term memory (LSTM)-based model to study latent driving policies reflected by the dataset.

\subsection{Research Objective}

In this paper, the problem is to learn an autonomous driving policy model from which the given Waymo Open Dataset is most likely to generate. The policy model generates a driving action given that the environment is in a certain motion state. The response refers to the instantaneous acceleration $\textbf{a}$ (including the magnitude and steering angle). In a realistic situation, $\textbf{a}$ can be generated by controlling the gas pedal and the brake. The environment is defined as measurable variables that describe the surroundings of an AV including 12 kinematics and distance measures. Those variables were collected by radars, lidars, cameras and other sensors installed on Waymo cars. There are various driving scenarios in the dataset, among which we only focus on one specific scenario: car-following on highways or urban roads.

\subsection{Contributions}

To the best of our knowledge, this paper is the first-of-its-kind to develop a driving model for existing level-5 autonomous cars using Waymo’s sensor data. Due to manufacturers' proprietary protection, no documentation has revealed how the existing AV fleets are actually programmed to drive and interact with other road users on public roads. 
Accordingly, academic researchers have to make various assumptions prior to controller design. 
Our work can be used as a general autonomous driving model for the transportation community on AV-related studies. 
Specifically, we have made the following contributions: 
\begin{itemize}
\item Design advanced LSTM-based policy models  with an encoder-decoder structure, which contains multiple stacked LSTM cells;
\item Evaluate the proposed LSTM-based policies by comparing to baselines; and
\item Investigate how using different feature sets as input affects the performance and training efficient of the models.
\end{itemize}

\section{Related Work}

\subsection{Behavioral Cloning}

Behavioral cloning is a popular approach to mimic the actions of human drivers. Mariusz Bojarski \textit{et al} \cite{BojarskiTDFFGJM16} from NVIDIA presented a convolutional neural network (CNN) to imitate driving behavior and predict steering commands. The trained neural network maps raw pixels from a single front-facing camera to basic steering commands. This system can successfully handle complicated scenarios, such as areas with blurred vision or on unpaved roads.

\subsection{Convolutional Neural Networks}

Convolutional Neural Network (CNN) is a widely applied mechanism in the computer vision field. In 2012, \citeauthor{NIPS2012_4824}\cite{NIPS2012_4824} proposed a novel deep learning architecture using CNNs and demonstrated its competence in the ImageNet\cite{5206848} contest. \citeauthor{Simonyan15} \cite{Simonyan15} proposed a deeper CNN network compared to AlexNet and proved that deeper models have better capacity and lower error rates. To develop deeper network models, \citeauthor{43022}\cite{43022} introduced the idea of inception, which improved the utilization of the computing resources inside the network. Moreover, \citeauthor{He_2016_CVPR} \cite{He_2016_CVPR} proposed the famous CNN network, ResNet, which further increased the depth of the network by adding residual blocks and reached a stunning depth of 152 layers. Later, the ResNet was refined by \citeauthor{10.1007/978-3-319-46493-0_38} \cite{10.1007/978-3-319-46493-0_38} by introducing identity mappings. 
\cite{liao2018large} have succesfully applied ResNet for taxi demand forecasting. 
In this work, the refined ResNet is treated as a feature extractor for camera images in Waymo Dataset.

\subsection{Recurrent Neural Networks}
Recurrent Neural Network (RNN) is advantageous in processing time sequence data as its capability of storing the information through the internal state to behave temporally and dynamically. LSTM network, the refined version of RNN, solves the problem of gradient vanishing or exploding arising in original RNNs\cite{DBLP:journals/corr/abs-1211-5063}. Thus, many researchers adopt LSTMs as their primary models on various time sequence tasks such as machine translation\cite{DBLP:journals/corr/ChoMGBSB14}, activity recognition\cite{8121994} and trajectory prediction\cite{Alahi_2016_CVPR}. 

\subsection{Trajectory Prediction}
In the problem of trajectory prediction, the future prediction is usually based on past experience or the history of previous trajectories. Thus, LSTMs, designed to process time sequence data, can likely perform well on such a task. \citeauthor{Alahi_2016_CVPR} \cite{Alahi_2016_CVPR} proposed a novel LSTM model in predicting human trajectories in the crowded space and \citeauthor{8354239} \cite{8354239} incorporated more information including occupancy map and camera image to build a robust model in predicting pedestrian trajectories. Such ideas are also effective in predicting cars' trajectories as cars and pedestrians share common motion patterns to some degree. Inspired by their work, we will consider using LSTM-based approaches to process the time series data in the Waymo dataset and learn the driving policy underlying it.

\section{Data}

Waymo Open Dataset is the largest, richest and most diverse AV datasets ever published for academic research \cite{sun2019scalability}. This dataset, collected from Waymo level-5 autnomous vehicles in various traffic conditions, comprise radar, lidar and camera data from 1000 20-second segments (as of December 2019) with labels. In this section, we introduce details about the Waymo dataset, as well as how the data is preprocessed before being fed into several machine learning models. 

\subsection{Labels}

Labels in this dataset refer to kinematics and spatial parameters of objects, which are represented as bounding boxes. Specifically, one kind of labels, \textit{type}, is classified into \textit{pedestrian}, \textit{vehicle}, \textit{unknown}, \textit{sign} and \textit{cyclist} categories. Detailed information is provided for each label, among which we especially pay attention to the coordinates of the bounding boxes, velocities $\textbf{v}$, and accelerations $\textbf{a}$ in the subsequent feature extraction step.

\subsection{Coordinate Systems}

Three coordinate systems are provided in this dataset: global frame, vehicle frame, and sensor frame. Some raw features are represented in unintended coordinate systems. In order to maintain consistency, it is crucial to transform data into the correct coordinate system. The dataset also provides vehicle pose $\textbf{VP}$, a $4 \times 4$ row matrix, to transform variables from one coordinate system to another. The formula below shows the most frequent transformation that is employed in our experiments \cite{waymo_data}. 

$$\textbf{X}_{VF} \cdot \textbf{VP} = \textbf{X}_{GF},$$
where $\textbf{X}_{VF}$ and $\textbf{X}_{GF}$ refer to the vehicle frame and the global frame respectively.

\subsection{Data Size}

According to the data format, 1000 segments are packed into multiple compressed files (tars) with a size of 25 GB each. In our experiments, 32 training tars are used as the training set and 8 validation tars are used as the testing set. The total number of videos extracted from the segments is 45000.

\subsection{Acceleration Computation}

Because one's instant acceleration of is not directly available in the dataset, the ``ground truth'' for training and evaluation needs to be computed by velocity differences. We designed the formula below to calculate the acceleration between two consecutive video frames, where the acceleration of the initial frame is set to zero.

$$
\textbf{a}_t = \left\{
    \begin{array}{ll}
      0, &t=0\\
      \frac{\textbf{v}_t - \textbf{v}_{t-1}}{\Delta t}, &t>0
    \end{array}
  \right.
$$
where $t$ is the frame index, $\textbf{a}_t$ and $\textbf{v}_t$ are the acceleration and velocity of the AV at the time frame $t$, respectively.

\begin{figure}[H]
\centering
\begin{minipage}{.45\textwidth}
  \centering
  \includegraphics[width=.9\linewidth]{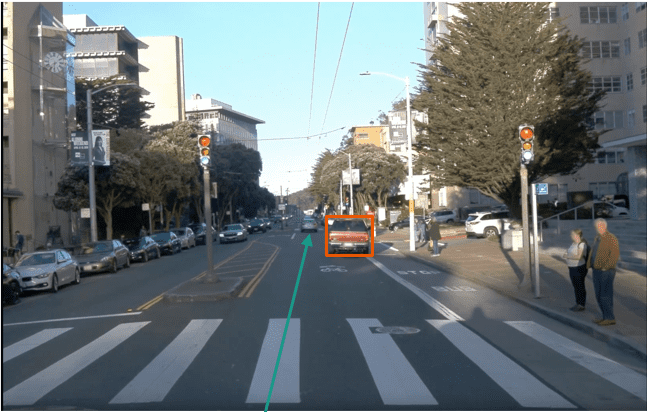}
  \\(a) Detection tolerance = 0
\end{minipage}
\begin{minipage}{.45\textwidth}
  \centering
  \includegraphics[width=.9\linewidth]{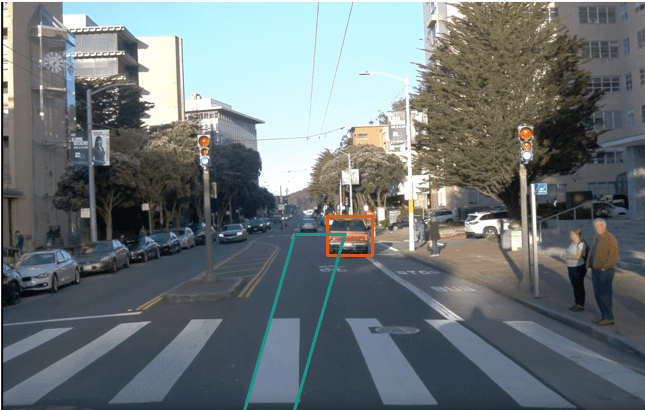}
  \\(b) Detection tolerance = 1
\end{minipage}
\caption{Examples of two different detection tolerances applied to the same scenario, displayed on visualizations of camera images in Waymo Open Dataset.}
\label{tol}
\end{figure}

\subsection{More Features}

In addition to one's velocity which is represented in the global frame, it is also beneficial to take the information of the front car into account while studying the change in the acceleration of the AV. As mentioned previously, the acceleration and velocity data are provided per bounding box in the vehicle frame, it is necessary to transform them into the global frame to keep consistency.

Furthermore, we use a simple geometrical procedure to detect the existence of the front car. As shown in Figure~\ref{tol}, the procedure check if a hypothetical ray (green) starting from the position of the AV intersects with the front car (the red bounding box). However, one defect is that there could be many misses by using a single ray. So we optimized this procedure by utilizing a rectangular ray to represent the AV’s view with some detection tolerance. As a result, the following quantities related are extracted as part of the features: velocity ($v_\text{fx}, v_\text{fy}, v_\text{fz}$) and acceleration of the front car ($a_\text{fx}, a_\text{fy}, a_\text{fz}$) and the relative distance to the front car ($\text{d}x, \text{d}y$). Additionally, enlarging the tolerance range may lead to detection mistakes when there are multiple vehicles ahead, thus the number of vehicles is also included to describe the complexity of the scenario, which is denoted as $num\_v\_labels$. For frames where the front car is not detected, the previously mentioned quantities are set to 0.

\subsection{Image embedding}

\begin{figure}[H]
\begin{subfigure}{.33\textwidth}
  \centering
  \includegraphics[width=.9\linewidth]{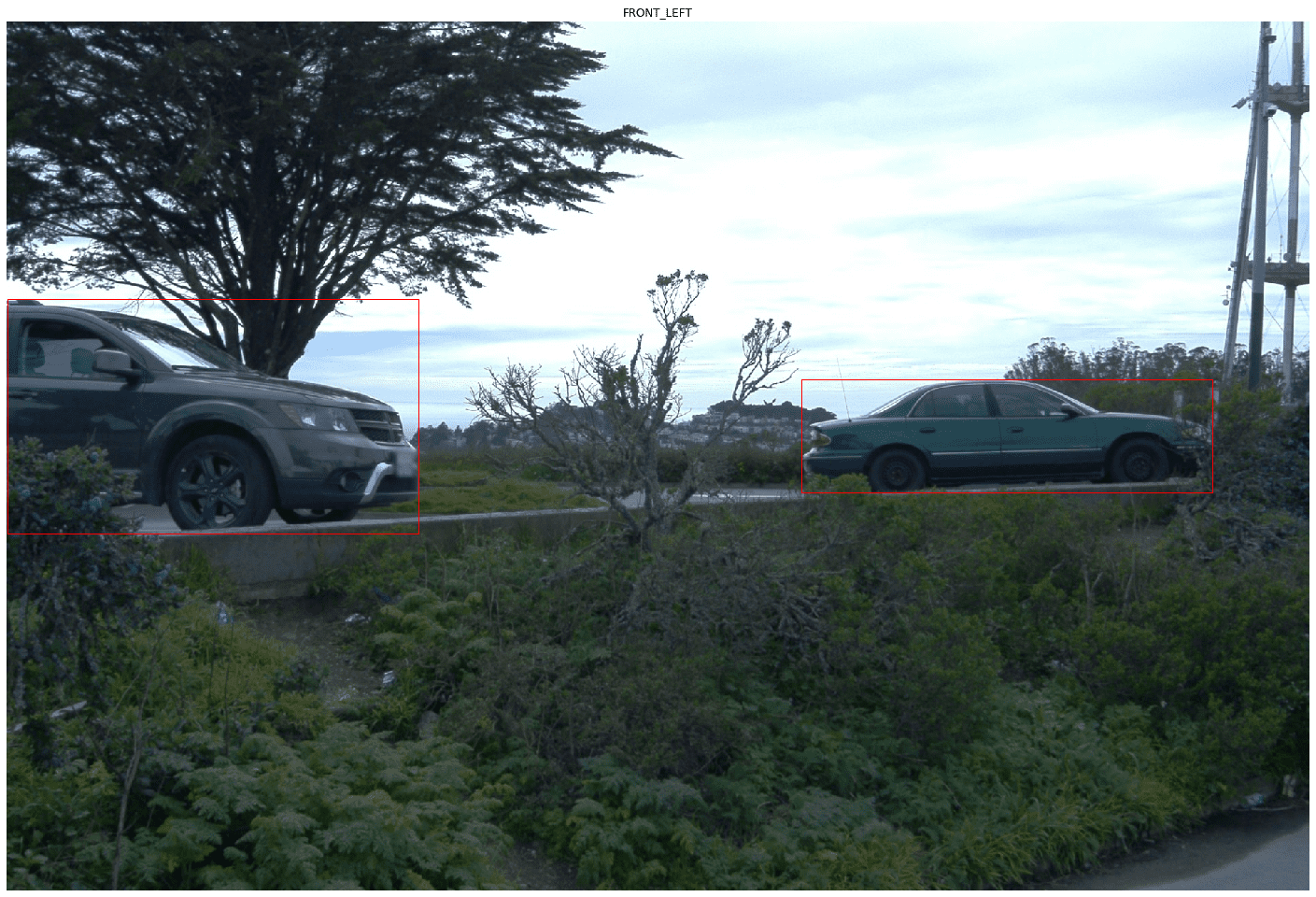}  
  \caption{Front-Left}
  \label{fig:sub-first}
\end{subfigure}
\begin{subfigure}{.33\textwidth}
  \centering
  \includegraphics[width=.9\linewidth]{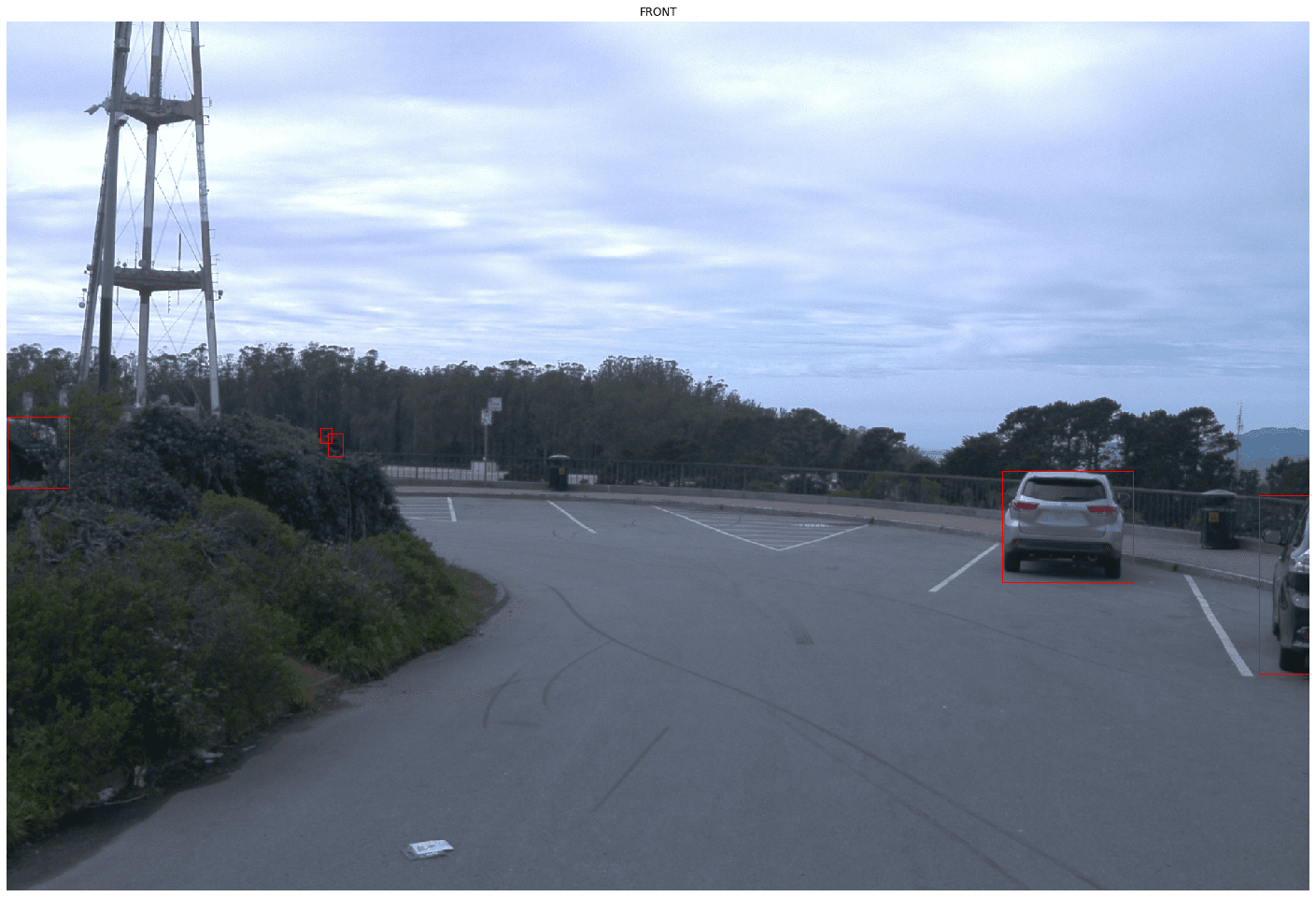}  
  \caption{Front}
  \label{fig:sub-second}
\end{subfigure}
\begin{subfigure}{.33\textwidth}
  \centering
  \includegraphics[width=.9\linewidth]{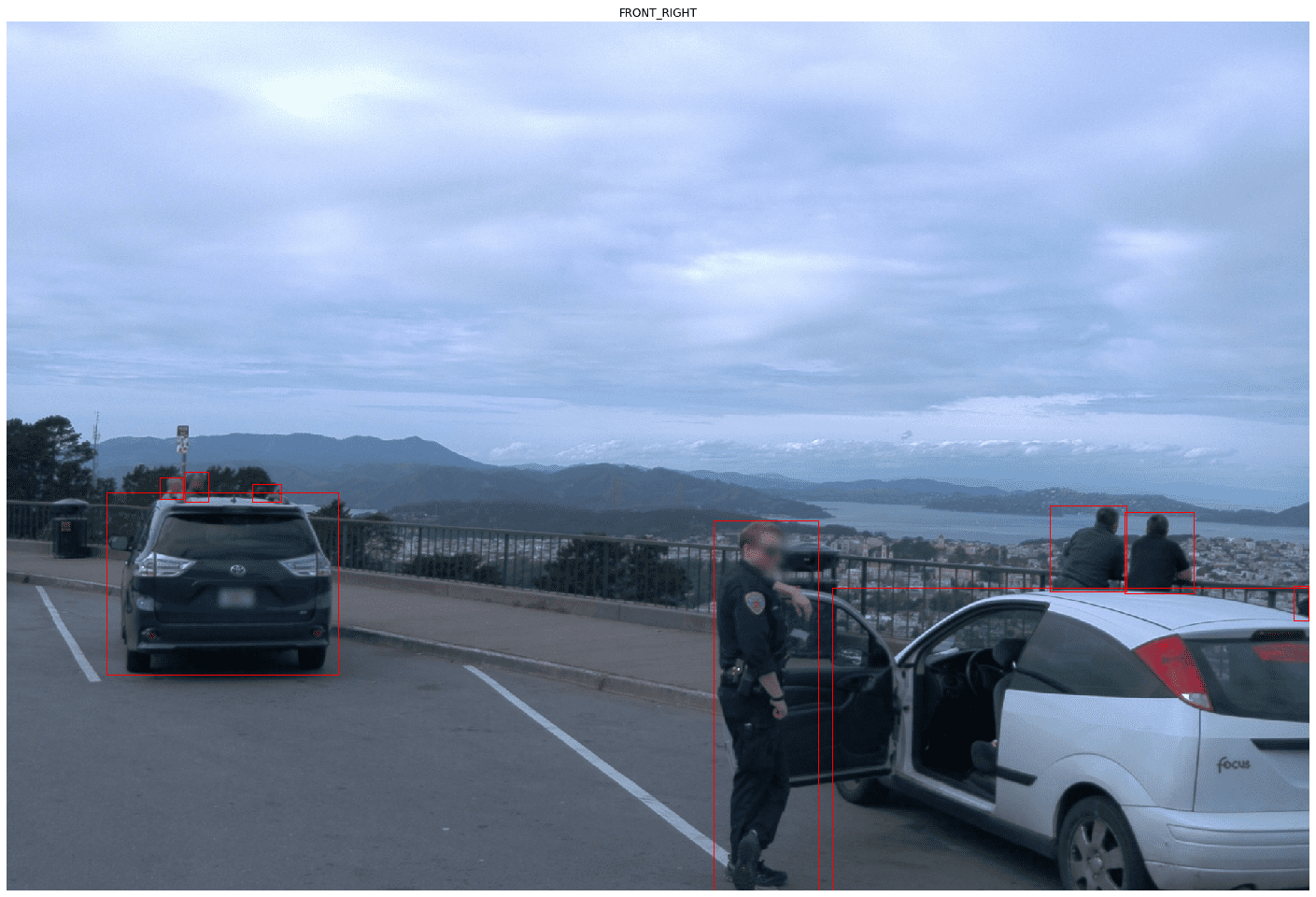}  
  \caption{Front-Right}
  \label{fig:sub-third}
\end{subfigure}
\newline
\vspace{1em}
\newline
\begin{subfigure}{.5\textwidth}
  \centering
  \includegraphics[width=.8\linewidth]{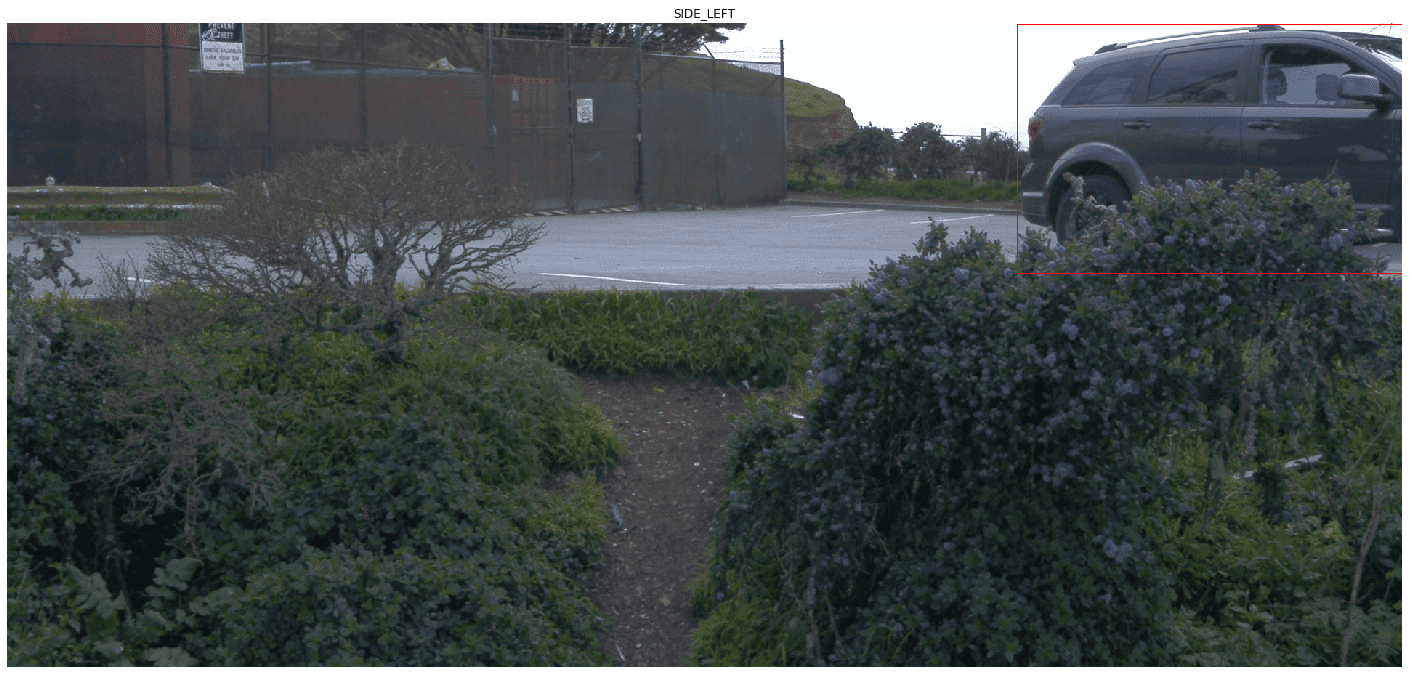}  
  \caption{Side-Left}
  \label{fig:sub-fourth}
\end{subfigure}
\begin{subfigure}{.5\textwidth}
  \centering
  \includegraphics[width=.8\linewidth]{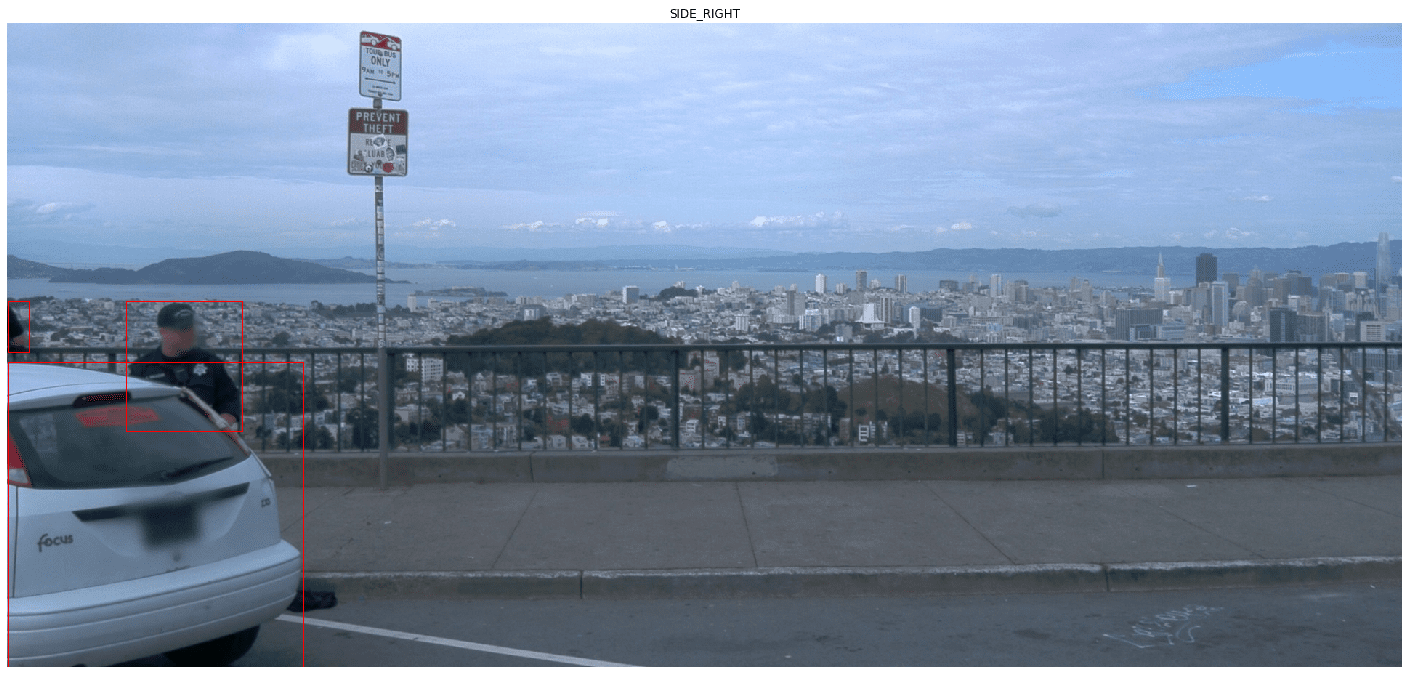}  
  \caption{Side-Right}
  \label{fig:sub-fifth}
\end{subfigure}
\caption{One example of images captured by cameras of five different views from segment-11004685739714500220 of tar training\_0000 in Waymo Open Dataset.}
\label{fig:five-images}
\end{figure}

As shown in Figure~\ref{fig:five-images}, there are five cameras installed on the AV, facing towards \textit{front, front-left, front-right, side-left, and side-left} respectively. These images reflect the time-series information of the moving vehicle with relatively smoother variation than numerical data, which helps to prevent spiky prediction between consecutive frames. Therefore, we utilize image embedding, a concise representation (vector) of images, as a part of input data to train the models. More details of image embeddings will discuss such embedding in the LSTM-based model sections.

\section{Baseline Models}

In order to establish benchmarks for comparison, we propose several baseline models using different methods including neural network models and decision-tree-based ensemble methods. In this section, we focus on the structures and principles of the baseline models.

\subsection{Neural Network Models}

\begin{figure}[h]
  \centering
  \includegraphics[width=\textwidth]{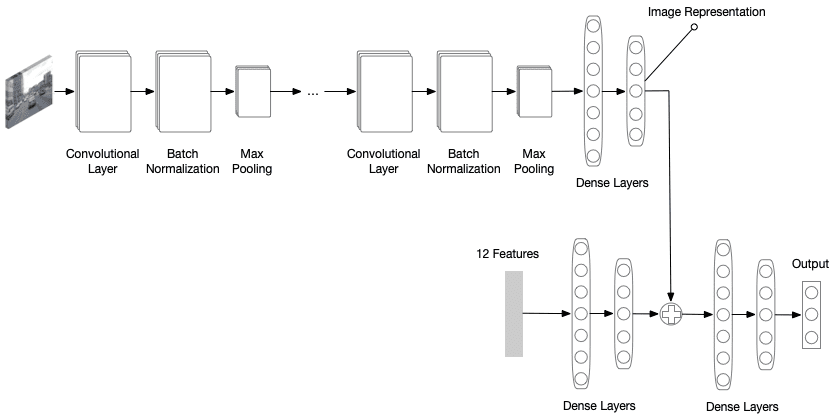}
  \caption{The composited model of NN and CNN. Outputs of dense layers are concatenated to form new dense layers which generate the final output. Notice that the output of the dense layer from the CNN network is actually the trained representation of the input image.}
  \label{cnn_nn}
\end{figure}

\begin{itemize}
    \item \textbf{NN}: Since the numeric and image data can both represent the surroundings and affect the decisions made by the AV, the first baseline model is a simple deep neural network that takes 12 features as input and owns 2 hidden layers with 8 and 4 neurons respectively. This model aims at capturing the characteristic of the 12 features, as well as the relationship between accelerations and the surrounding environment.
    
    \item \textbf{CNN}: The second model is a traditional convolutional neural network with convolutional, batch normalization, and pooling layers. Grid search is applied to determine the optimal parameters during training. Video images of each frame in the dataset are used as inputs to predict the acceleration of the same frame. Then, the output of the multiple layers is fed into dense layers to extract feature embedding of the image and generate the output.

    \item \textbf{NN + CNN}: To better represent a frame using the image embedding as well as the 12 features corresponding to a frame, the third model combines the NN and CNN models. Specifically, as shown in Figure~\ref{cnn_nn}, the last dense layers of both models are concatenated to new dense layers which eventually generate the final output. 
\end{itemize}

\subsection{Ensemble Methods}

Another popular category for regression problems is the ensemble methods. For a thorough exploration of the dataset, we attempted several ensemble methods and compare their prediction results.

\begin{itemize}
    \item \textbf{XGBoost}: XGBoost is a decision-tree-based ensemble Machine Learning algorithm that is built on a gradient boosting framework. It uses a more regularized model formalization to control over-fitting, which frequently produces superior performance in competitions. 
    
    \item \textbf{Light Gradient Boosting}: Another popular boosting algorithm, Light Gradient Boosting \cite{NIPS2017_6907}, uses the gradient-based one-side sampling for splitting, instead of growing trees horizontally (increasing levels), it grows vertically to increase the number of leaves, which minimizes loss to a higher degree than growing by levels.
    
    \item \textbf{Stacked Linear Regressor}: Stacked linear regressor \cite{Breiman_1996} is a more comprehensive ensemble method that linearly combines different models. Unlike the other boosting regressors, it uses a second-layer learning algorithm, where each model in the second layer finds its optimum. Moreover, the second-layer regressor can be built by more complex models, which overcomes the limitations of the single linear regression model. 
\end{itemize}

\section{LSTM-based Model}

\subsection{Basic Model with 12 Features}
\begin{figure}[H]
  \centering
  \includegraphics[width=1\textwidth]{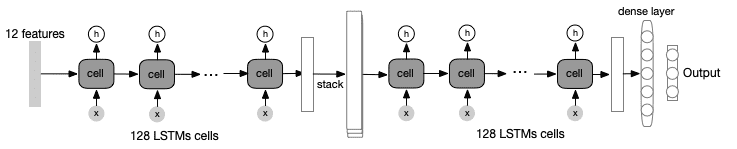}
  \caption{Given one video clip with a frame-length of 10, the input is the vector consists of 12 features from these 10 frames. The output is the acceleration for the next 5 frames starting from the end of the video clip. The "encoder" module contains 128 LSTM-cells and the "decoder" module contains 128 LSTM-cells}
  \label{architecture_basic}
\end{figure}

One of the straightforward ways to build the acceleration prediction model is to treat 12 basic features as the input of the model. The "encoder-decoder" architecture proposed for  trajectory prediction in SS-LSTM \cite{8354239} is a suitable architecture for the acceleration prediction problem as the acceleration curve is a trajectory based on past experiences. While SS-LSTM uses different input information such as the occupancy map, the 12 basic features are packed into one single input channel in our proposed basic model. These features are fed into an "encoder" module and this "encoder" extracts the key information from input features and generates an intermediate result, or more precisely, a latent representation of these features. The intermediate result is then forwarded into a "decoder" module, which decodes the representation and outputs the acceleration prediction. The figure of this architecture is shown as Figure~\ref{architecture_basic}.

However, the limitation of information provided by the 12 input features cannot be ignored. It is difficult for the model to capture and analyze the complex surrounding environment for the AV by these simple and limited numerical features, and the model trained on these features may fail to give a satisfying prediction. High-quality features with more key information are needed to improve the model capacity.

\subsection{Advanced Model with Image Inputs}
Given the limitation of the basic model, more features from the Waymo Open Dataset should be incorporated and one of the most critical data that can be used is the camera image, which perfectly records the changing environment of the AV. It is obvious that a raw camera image cannot be fed into the basic model directly as an image is a two-dimensional data point while the input is required to be a one-dimensional vector. Thus, converting the raw image into a vector whose elements represent the key content of the image is necessary. 

Naturally, CNN is a popular and excellent image feature extractor that converts an image into a flatten vector and adding one CNN in front of the "encoder" module in the basic model seems to be reasonable. Training the CNN module with the "encoder" module and the "decoder" module as a whole can yield an end-to-end advanced model, much more complex than the basic model. But training this model from scratch can be time-consuming because of the complexity of a deep CNN module. One quick but efficient solution is to use a pre-trained CNN module to convert the raw image into a vector without any fine-tuning, as the existing popular pre-trained CNN networks are mostly well trained on massive datasets and should generalize well. One of the best CNN networks is the ResNetV2 \cite{10.1007/978-3-319-46493-0_38} and the output of its second last layer can be a high-quality representation of the input image. Given such an image feature extractor, the front camera image in the Waymo Open Dataset is chosen as the major input as this camera should obtain the most critical environment of the AV and contain key information to determine the future acceleration, or the car behavior decision.

\begin{figure}[h]
  \centering
  \includegraphics[width=1\textwidth]{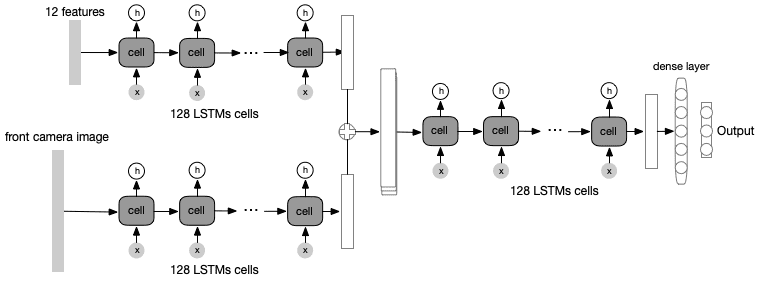}
  \caption{Noticed that the "image" input is actually a vector containing key image content. The first channel input is the 12 features for the observed 10 frames. The second channel input is the 1576-dimensional representation of front camera images from these 10 frames. Such representation is extracted from the second last output of a pre-trained Resnet152v2. The output is the acceleration for the future 5 frames. The "encoder" module contains 128 LSTM-cells and the "decoder" module contains 128 LSTM-cells}
  \label{architecture_image_front}
\end{figure}

The architecture of such an advanced model is similar to the previous basic model. An "encoder-decoder" structure is maintained to learn the information hidden in the input features. The difference is that the front camera images are treated as additional inputs. The detail of this architecture is shown as Figure~\ref{architecture_image_front}.

\begin{figure}[t!]
  \centering
  \includegraphics[width=1\textwidth]{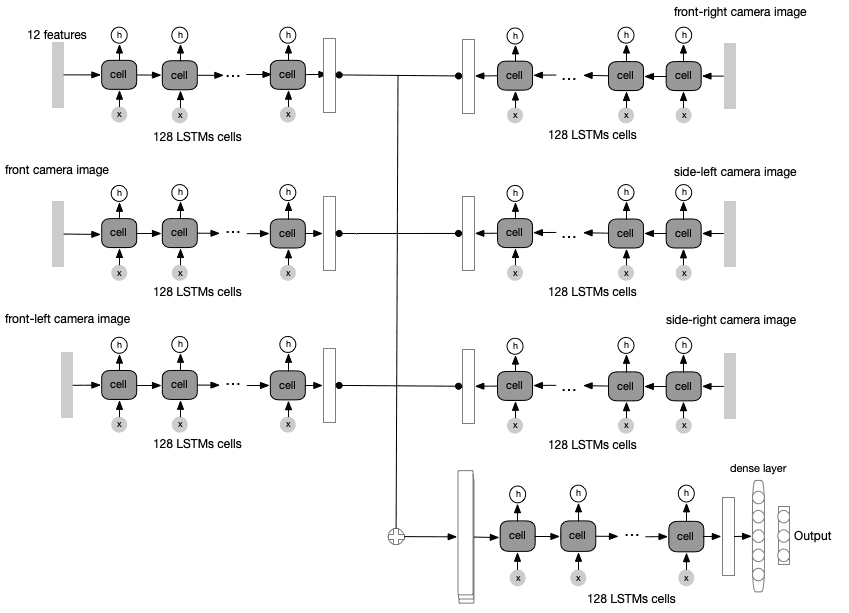}
  \caption{The first channel input is the 12 features for the observed 10 frames. The rest of the input is the 1576-dimensional representation of camera image from different views in the observed 10 frames. The output is the acceleration for the future 5 frames. All "encoder" modules contain 128 LSTM-cells and all "decoder" modules contain 128 LSTM-cells}
  \label{architecture_image_all}
\end{figure}

Adding more cameras from different views may help the model even understanding the driving environment better as adding a front camera has improved the model performance to a significant degree. In the Waymo Open Dataset, cameras from five different views are provided, which are the front camera, the front-left camera, the front-right camera, the side-left camera, and the side-right camera respectively. Incorporating all these five cameras can obtain a more advanced model than the previously trained models. The module architecture of the model is similar and the only change is to add 4 extra input channels for the rest of four cameras besides the front camera. The architecture of this model is shown as Figure~\ref{architecture_image_all}

\section{Experiments and Results}

Each model is evaluated on the same test data sampled from Waymo Open Dataset, which is called the validation part of the dataset. To determine the accuracy of predicting acceleration from the trained model, Mean Absolute Error (MAE) is chosen as the major metric. The MAE value is calculated by taking the average on all MAE errors of the video clips from the test data. VM instance from Google Cloud Platform is used to train all models mentioned and the VM has one single NVIDIA Tesla K80. 

\subsection{Hyper-Parameter}
In the experiments, the major hyper-parameters are the following:
\begin{itemize}
    \item \textbf{The length of the previous frames} This decides how much the model can see in the history. If the length is too short, the model may perform poorly as it could not obtain enough history to predict accurately. If the length is too long, the model is too complex and waste resources. We set this number to 10.
    \item \textbf{The length of the future frames} This decides how much the model needs to predict for the future. If the length is too long, the model is too complex and needs more history to support good prediction. We set this number to 5.
    \item \textbf{training epochs} We set this to 300 in order to obtain a well-trained model.
\end{itemize}

\subsection{Results}

From Table~\ref{baseline_results}, the model using 12 features and the front camera image has outperformed most of the other models and it is proved that adding image has pushed the model performance to another level compared with the one only using the 12 basic features. However, the model with all cameras has similar performance with the model using only the front camera, which is not consistent with the previous assumption. The reason why the five-camera model failed to perform even better is that the rest of the camera image is noisy input for the model and may likely interfere with information from the front camera, resulting in a slight decrease of the model performance.

\begin{table}[H]
\caption{MAE comparison. The minimum loss is bolded.}
\label{baseline_results}
\centering
\begin{tabular}{ccccc}
\toprule
\textbf{Models} & \textbf{MAE X} & \textbf{MAE Y}\\
\midrule
NN & 0.4014 & 0.4312\\
CNN & 0.3272 & 0.3123\\
NN+CNN & 0.2985 & 0.2802\\
XGBoost & 0.3502 & 0.3537\\
Light Gradient Boosting & 0.3459 & 0.3444\\
Stacked Linear Regressor & 0.3623 & 0.3222\\
\midrule
LSTM with 12 features & 0.3179&0.3088\\
LSTM with front camera & 0.1379 & \textbf{0.1278}\\
LSTM with all cameras & \textbf{0.1327} &0.1363 \\
\bottomrule
\end{tabular}
\end{table}

\subsection{Training Efficiency}
The time efficiency of different models is also evaluated during the experiments. From  Table~\ref{table:time}, it is obvious that adding images will lead to the rapid growth of training time due to the increase of the model complexity. When all five camera images are introduced into the input data, training takes much more time than only using front camera images. But the previous result has shown that the difference between the performance of the model using only the front camera and that using all cameras is small. Thus, it is more reasonable to only use the front camera image as the image input because it is more computationally efficient than using images from all five cameras.
\begin{table}[H]
\caption{Approximate training time of different models.}
\centering
\begin{tabular}{ccc}
\toprule
\textbf{Model}	& \textbf{Approximated GPU Training Time}\\
\midrule
LSTM with 12 features & 1.8 hour\\
LSTM with front camera  & 9.0 hours\\
LSTM with all cameras &  37.8 hours\\
\bottomrule
\end{tabular}
\label{table:time}
\end{table}

\subsection{Visualization}

\begin{figure}[H]
\begin{subfigure}{\linewidth}
  \centering
  \includegraphics[width=.9\linewidth]{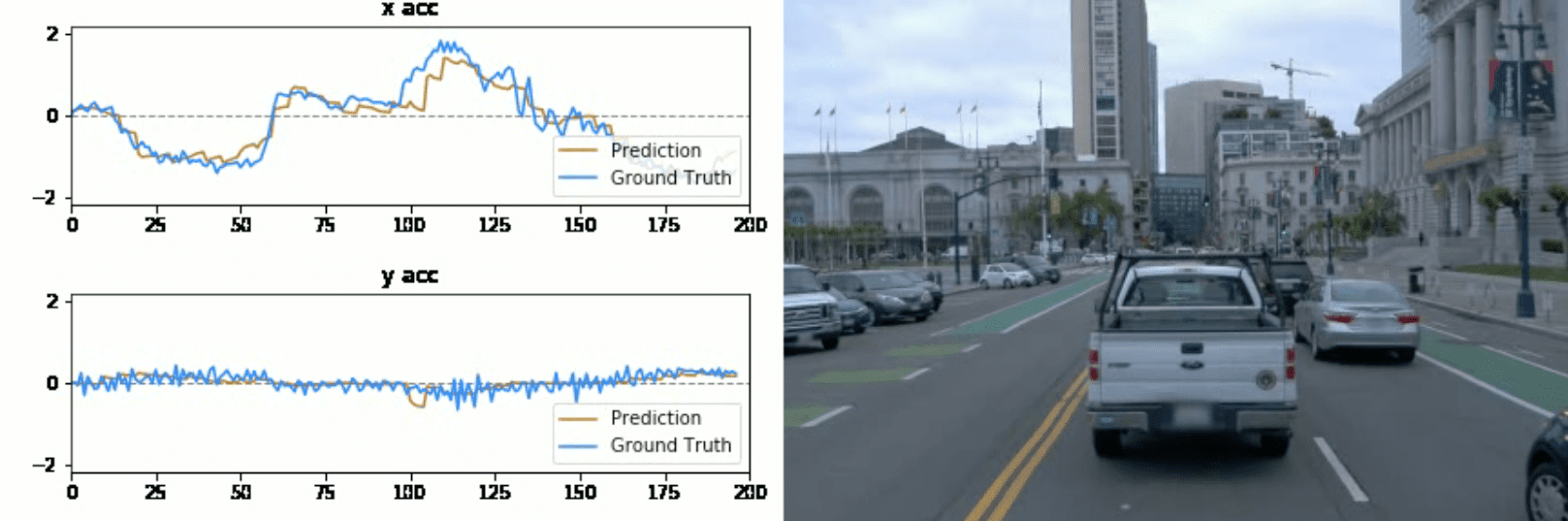}  
  \caption{The last frame of segment-10289507859301986274 from tar validation\_0001 and corresponding prediction.}
  \label{fig:vis1}
\end{subfigure}
\newline
\vspace{1em}
\newline
\begin{subfigure}{\linewidth}
  \centering
  \includegraphics[width=.9\linewidth]{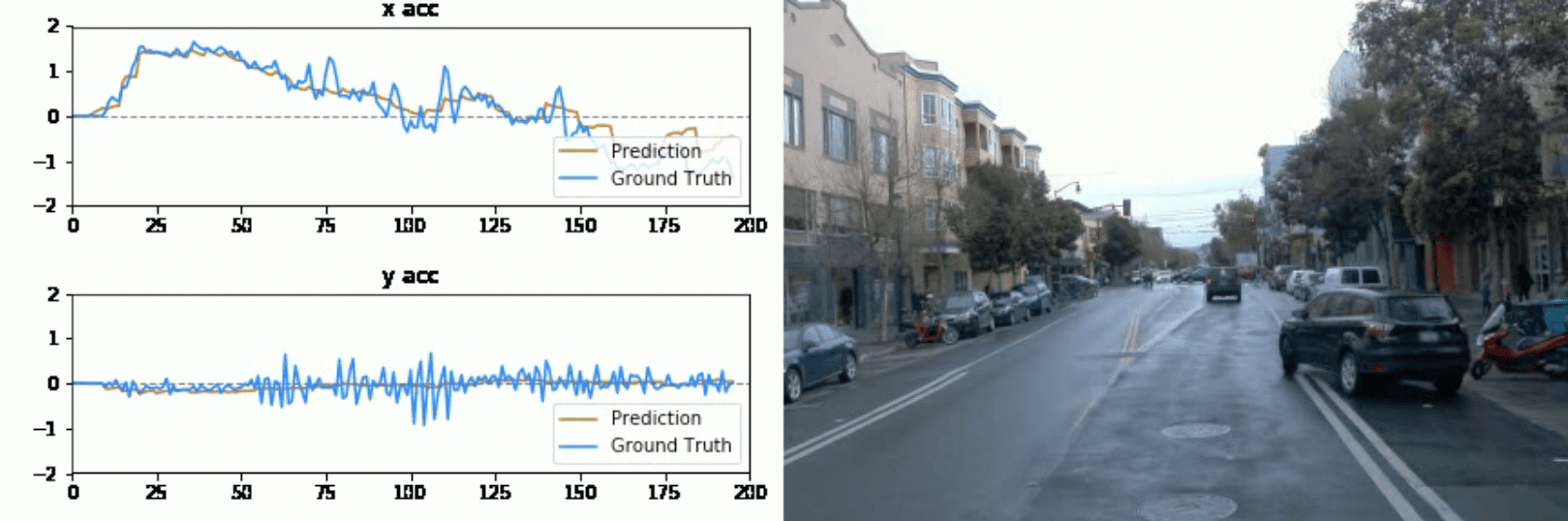}  
  \caption{The last frame of segment-12496433400137459534 from tar validation\_0001 and corresponding prediction.}
  \label{fig:vis3}
\end{subfigure}
\newline
\vspace{1em}
\newline
\begin{subfigure}{\linewidth}
  \centering
  \includegraphics[width=.9\linewidth]{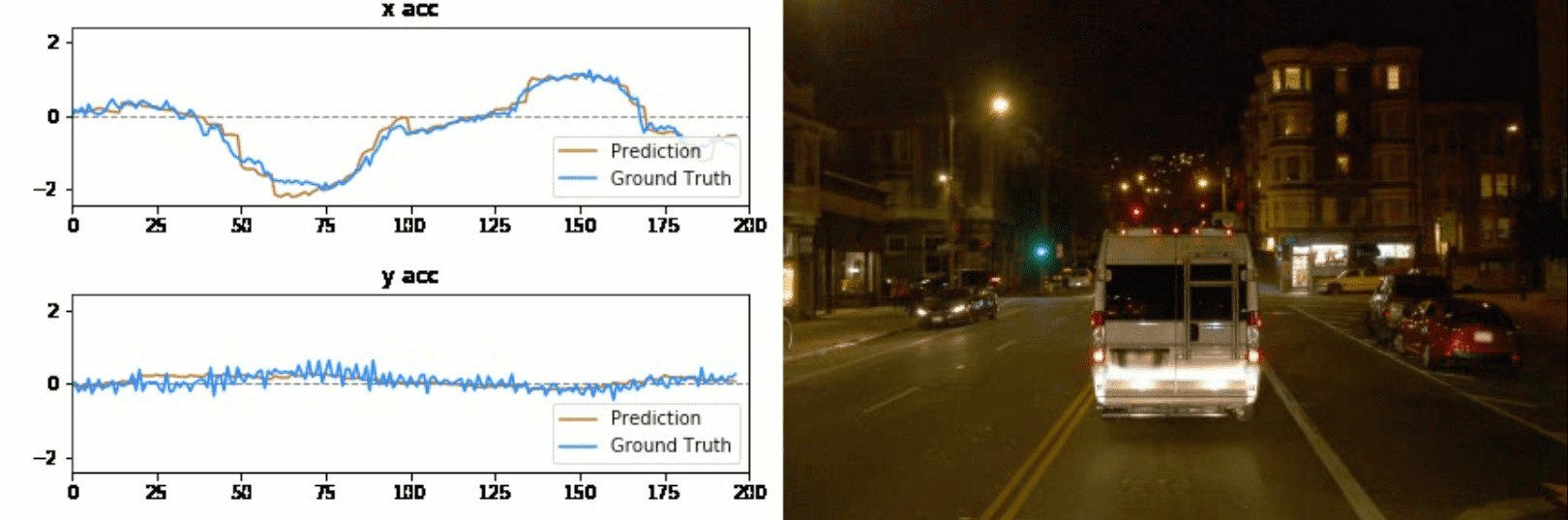}  
  \caption{The last frame of segment-11660186733224028707 from tar validation\_0001 and corresponding prediction.}
  \label{fig:vis2}
\end{subfigure}
\caption{Model prediction with synchronized frames captured by the front-camera on the right.}
\label{fig:vis-images}
\end{figure}

To visually inspect how well our model performs, we develop a visualization tool, which takes the predicted accelerations and raw camera images as the input, and generates a video as the output. 
In the three data plots of Figure~\ref{fig:vis-images}, the horizontal axis represents the frame index of the segment, while the vertical axis means the acceleration value. The ranges of the vertical axes are all limited to $[-2.0, 2.0]$ to prevent misinterpretations.
The generated video shows both predicted and ground-truth values for longitudinal and lateral accelerations (marked as ``\textit{x acc}'' and ``\textit{y acc}'', respectively). 
Combining the data plot (left) and actual video frames (right) give us an intuitive way to inspect prediction deviation within a continuous spatial and temporal range.

Generally, 
our model maintains a decent performance in more complex scenarios such as nighttime (as in Figure~\ref{fig:vis2}). 
The visualization results show that our predicted accelerations behave more cautiously than Waymo cars. As indicated in Figure~\ref{fig:vis1}, 
our model accelerates and decelerates more gentle than Waymo's model. Also, because of the smoothing step in preprocessing, our model generates a smoother acceleration curve than Waymo's, as shown in Figure~\ref{fig:vis3}.

\section{Conclusions and future work}

From the experimental results, it is observed that the LSTM-based model has better performance than the baseline models, which also supports our assumption that the acceleration prediction is a time-series problem. Inspired by the architecture of LSTM, the "encoder-decoder" structure is adopted in all LSTMs-based models. At first, only the basic numeric features such as velocity and distance between the front car and the current AV are used to train the model. Such an LSTM-based model has already outperformed the baseline models without using CNN or those considering time-dependency in the data. To further increase the model performance, camera images are introduced to help the model output a more accurate result. The result shows that the LSTM-based model with image has significant improvement. However, the model using the front camera only shares similar performance with the model using all five cameras while the latter took much longer time to train. Therefore, we conclude that the LSTM-based model with the front camera images is an effective and efficient model for acceleration prediction.

This work can be extended in several ways. 
One limitation of the experiments we carried out is that in the current training and testing process, the ``online approach'' takes the ground truth relative distance and velocities as input to the network after predicting the first frame after the initial 10 frames. However, in the scenario of the real world, the input should be taken from (or accumulated from) the prediction of previous frames. Therefore, our current approach may cause the loss to accumulate continually. To solve this problem, our future work is to generate predictions of one frame immediately after the initial ten frames, and then compute the loss between that specific frame and the ground truth data. Another future direction is to train a more generalizable model for other traffic scenarios than car-following, for example, lane-changing, merging and diverging.

\acknowledgments{We would like to express our great gratitude to other collaborators who have contributed to this work. Thank Qian Zheng, Gerardo Antonio Lopez Ruiz, and Yumeng Jiang for featuring engineering at an early stage, coming up with the definitions of the crucial ``12 features'' on which the models were trained on, as well as developing and testing several baseline models such as the XGBoost method and so on. We also appreciate explorations made by Aashish Kumar Misraa, Naman Jain, and Saurav Singh Dhakad about including camera images into the training data, which greatly broadened our scope of mind. Also, thank Siddhant Gada and Aditya Das for implementing other neural network baseline models for comparison purposes. }

\bibliographystyle{elsarticle-harv} 
\biboptions{authoryear}
\bibliography{references}

\end{document}